\documentclass[review]{elsarticle}
\usepackage{setspace}
\usepackage{newtxmath}
\usepackage{adjustbox}
\usepackage{array}
\usepackage{booktabs} 
\usepackage[usenames,dvipsnames]{color}
\usepackage{float}
\usepackage{algorithm}
\usepackage{algpseudocode}
\usepackage{caption}
\usepackage{xr}
\usepackage{multirow}
\usepackage[numbers,sort&compress]{natbib}

\biboptions{numbers,sort&compress}
\journal{Journal of Manufacturing Processes}

\title{Adaptive few-shot learning for robust part quality classification in two-photon lithography}

\author[add1]{Sixian Jia} 
\ead{sixian@umich.edu}

\author[add1]{Ruo-Syuan Mei} 
\ead{rsmei@umich.edu}

\author[add1,add2]{Chenhui Shao\corref{mycorrespondingauthor}}
\cortext[mycorrespondingauthor]{Corresponding author: Chenhui Shao (chshao@umich.edu).}
\ead{chshao@umich.edu}

\address[add1]{Department of Mechanical Engineering, University of Michigan, Ann Arbor, MI 48109, United States}

\begin{document}
\begin{abstract}

Two-photon lithography (TPL) is an advanced additive manufacturing (AM) technique for fabricating high-precision micro-structures. While computer vision (CV) is proofed for automated quality control, existing models are often static, rendering them ineffective in dynamic manufacturing environments. These models typically cannot detect new, unseen defect classes, be efficiently updated from scarce data, or adapt to new part geometries. To address this gap, this paper presents an adaptive CV framework for the entire life-cycle of quality model maintenance. The proposed framework is built upon a same, scale-robust backbone model and integrates three key methodologies: (1) a statistical hypothesis testing framework based on Linear Discriminant Analysis (LDA) for novelty detection, (2) a two-stage, rehearsal-based strategy for few-shot incremental learning, and (3) a few-shot Domain-Adversarial Neural Network (DANN) for few-shot domain adaptation. The framework was evaluated on a TPL dataset featuring hemisphere as source domain and cube as target domain structures, with each domain categorized into good, minor damaged, and damaged quality classes. The hypothesis testing method successfully identified new class batches with 99-100\% accuracy. The incremental learning method integrated a new class to 92\% accuracy using only K=20 samples. The domain adaptation model bridged the severe domain gap, achieving 96.19\% accuracy on the target domain using only K=5 shots. These results demonstrate a robust and data-efficient solution for deploying and maintaining CV models in evolving production scenarios. 

\end{abstract}
\maketitle

\section{Introduction}\label{sec:intro}
Two-photon lithography (TPL), also known as direct laser writing, is an advanced additive manufacturing (AM) technology capable of fabricating highly precise, three-dimensional micro- and nano-scale structures from photo-reactive polymers \cite{marschner2023methodology,sun2024automated,oguguo2025characterization}. Unlike conventional lithography, TPL utilizes a non-linear two-photon absorption process, which occurs only at the precise focal point of a femtosecond laser \cite{zhang2020design,matining2025advances, jia2023physics}. This voxel-by-voxel fabrication approach enables the creation of complex 3D geometries with sub-micron resolution, finding significant applications \cite{hu2025multiphoton} in fields such as photonic crystals \cite{wang2023two,jee2021combining}, micromachines \cite{maruo2003force,lin2018microstructures}, and tissue engineering scaffolds \cite{lu2023fabrication,sun2025emerging}.

The remarkable potential of TPL across these diverse applications is, however, critically dependent on achieving both high structural part quality and geometric accuracy \cite{jia2025physics,dong2024filtered}, which a major challenge preventing widespread industrial adoption \cite{behera2021current, nasrin2024application, mei2024deep}. Traditionally, quality evaluation has relied heavily on manual, post-fabrication inspection using Scanning Electron Microscopy (SEM) \cite{wang2024two,sun2023situ}, which is time consuming, labor intensive, and cannot be integrated into the fabrication process, creating a significant bottleneck that is unsuitable for scalable manufacturing.

To address this bottleneck, computer vision(CV)-based approaches utilizing deep learning have shown as a promising solution for automated quality inspection \cite{islam2024deep,mei2025synthetic}. Several recent studies have validated the use of deep learning for automated quality classification using in-situ video data. Lee et al. \cite{lee2020automated}, for example, developed a deep learning model to detect part quality by analyzing video frames captured during the fabrication process. Building upon this concept, other researchers have explored similar in-situ monitoring systems. Hu et al. proposed a dual visual inspection system to optimize printing parameters and identify defects \cite{hu2024dual}. More recently, advanced architectures such as video transformers have been applied and demonstrated strong performance in automatically detecting part quality from video, which further establishes the potential of video-based analysis \cite{xiao2025video}.

While these video-based methods show promise, they face several major challenges that limit their practical applicability. First, the large volume of video streams requires significant storage and computational resources \cite{mohammadi2018deep, sahoo2022smart}. This large data volume, combined with the computationally intensive models required for analysis often leads to substantial latencies that are incompatible with inline monitoring \cite{nayak2021comprehensive,murthy2020investigations}. Second, deep learning models are dependent on large-scale, well-annotated datasets for effective training \cite{mei2025hybrid}. The process of creating and annotating such datasets for TPL are both time-consuming and expensive \cite{mu2019review}.

To mitigate these challenges associated, alternative CV frameworks based on static, post-fabrication images have been proposed \cite{jia2025end}. Our previous work, for instance, introduced an end-to-end framework combining segmentation with a Convolutional Neural Network (CNN)-based classification module \cite{jia2025end}. By using semi-supervised and unsupervised learning, the framework could effectively address data scarcity and achieve high classification accuracy \cite{jia2025end}. However, this framework was designed for a fixed set of defect categories and a single part geometry. It did not address the critical challenges of adapting the model when new, unseen defect classes are discovered, nor did it provide a mechanism for transferring knowledge to new part designs, which often exhibit significant domain shift.

To the best of our knowledge, the challenges associated with model adaptability addressing new, unforeseen defect classes and new part geometries have not yet been investigated within the TPL. However, a few recent studies have explored related topics for macro-scale AM processes. Such research generally falls into two main categories. The first category focuses on incremental learning methods, which are developed to efficiently update models with new defect classes without requiring complete retraining. These approaches have been applied, for instance, to tasks like 3D printer fault identification using fine-tuning \cite{zhao20253d}. The second category utilizes domain adaptation techniques to generalize models across different part designs, machine conditions, or product variants. Such studies have, for example, used deep learning to handle variant part geometries in layerwise imaging profiles \cite{imani2019deep}, employed domain-adversarial transfer learning for defect detection on new product variants \cite{tang2024domain}, and investigated adaptation to enhance digital twin reusability for monitoring systems \cite{xie2024investigation}. While adaptive monitoring and learning strategies have been explored in broader AM contexts, TPL’s voxel-scale fabrication, optical exposure dynamics, and microscopy-based inspection create a different regime that requires process and data-specific adaptations.

To address this research gap, this paper presents an adaptive CV framework designed to create, maintain, and deploy robust TPL quality classification models. The framework integrates key methodologies to address novelty, data scarcity, and domain shift, all built upon a scale-robust ResNet-18 and Spatial Pyramid Pooling (SPP) backbone. The first component is a statistical hypothesis testing framework based on  Linear Discriminant Analysis(LDA), which achieves 99-100\% accuracy in detecting batches of a new class. The second component is a two-stage, rehearsal-based incremental learning strategy that successfully integrates a new class from only $K=20$ labeled examples, achieving 92\% accuracy while mitigating catastrophic forgetting. The third component employs a few-shot Domain Adversarial Neural Networks (DANN) to bridge the significant domain gap between part geometries, yielding an accuracy of 96.19\% on the target domain with only 5 shots per class.

The remainder of this paper is organized as follows.  Section \ref{sec:method} introduces the proposed methodology for the backbone model, hypothesis testing, incremental learning, and domain adaptation. Section \ref{sec:dataset} details our dataset, highlighting the class distributions and the domain gap. The experimental results and discussions for each method are presented in Section \ref{sec:results}. Finally, Section \ref{sec:conclusion} concludes the paper and outlines potential directions for future research.

\section{Methodology} \label{sec:method}

\subsection{Backbone Model}\label{subsec:backbone}
For our feature extraction backbone, we utilize a ResNet-18 architecture, pre-trained on the ImageNet dataset \cite{he2016deep}. We adapt this standard model by truncating it before its final global average pooling and fully connected layers, thereby retaining the deep convolutional feature maps.

A key challenge in this task is that variations in structure size should not be confused with part quality. To ensure the model focuses on localized defect patterns rather than global part scale, we must aggregate features in a scale robust way. Therefore, to aggregate these features and capture multi-scale spatial information, the backbone is followed by a SPP module \cite{he2015spatial}. Our SPP layer is configured with pyramid levels of [4, 2, 1], which partition the feature map into 4x4, 2x2, and 1x1 spatial bins. Adaptive average pooling is applied within each bin, and the resulting 21 pooled vectors are concatenated.

\begin{figure}[H]
\centering
\makebox[\textwidth][c]{
\includegraphics[width=1\textwidth]{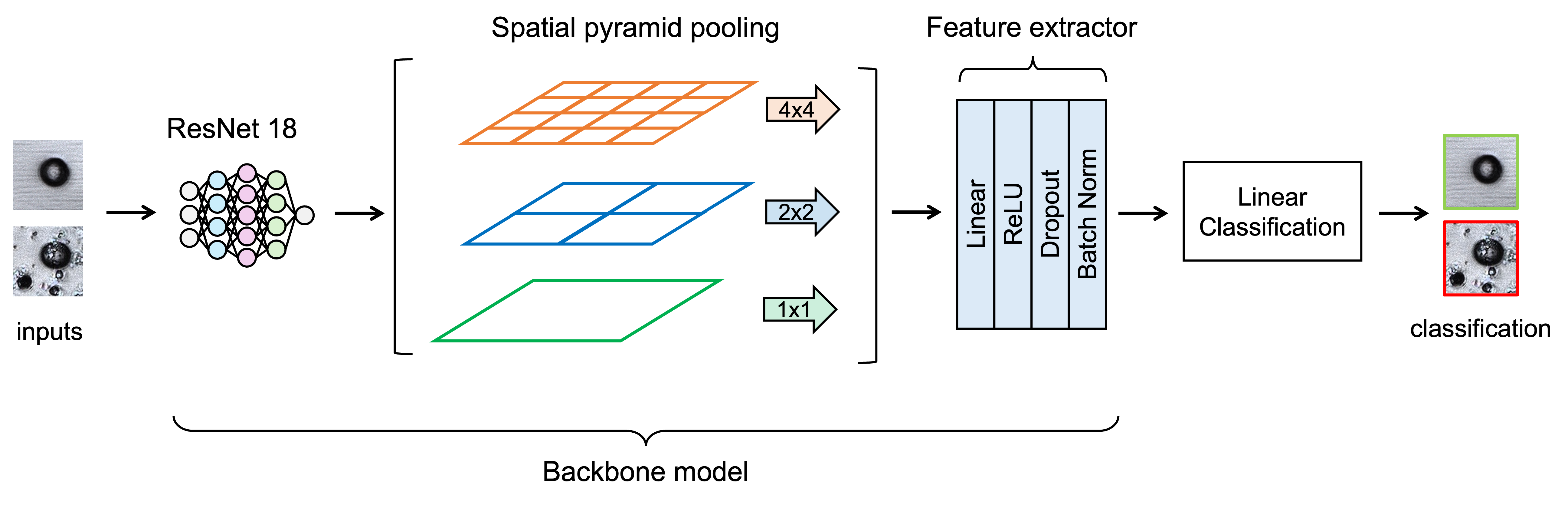}
}
\caption{Overview of the proposed backbone model.}
\label{fig:backbone}
\end{figure}

This multi-scale representation is then passed through a dedicated feature processor block, which consists of a fully connected layer projecting to 512 dimensions, a ReLU activation, a dropout layer, and a 1D batch normalization layer. Finally, a linear classifier layer maps these processed features to the output classes. This complete architecture, which we refer to as our backbone model as illustrated in Figure \ref{fig:backbone}.

For baseline training, we first form a 2-class problem using the two known defect states, while the third class is held out for later novelty/transfer evaluation. We employ a dual-view augmentation strategy: for each training image we generate two differently transformed views $(v_1, v_2)$ and feed them through the shared backbone with SPP head to obtain feature embeddings $(z_1, z_2)$. The main supervision is the standard cross entropy loss $\mathcal{L}_{CE}$ on the 2 class labels. To make the representation robust to view and scale changes, we add a scale consistency loss (SCL) defined as the mean squared error (MSE) between L2-normalized features:
\[
\mathcal{L}_{\text{SCL}} = \mathcal{L}_{\text{MSE}}\left(
\frac{z_1}{\|z_1\|_2},
\frac{z_2}{\|z_2\|_2}
\right).
\]
The final objective is
\[
\mathcal{L}_{\text{total}} = \mathcal{L}_{CE} + \lambda \mathcal{L}_{\text{SCL}},
\]
with $\lambda=0.1$ in all experiments. This auxiliary term specifically targets scale and view robustness, complementing the SPP based multi-scale aggregation in the backbone.

\subsection{Hypothesis Testing}
To validate the separability of new classes, we design a statistical hypothesis testing approach. This approach is intended to determine if a batch of new, unlabeled samples belongs to one of the known classes or to a new, unseen class.

In this experiment, we evaluate the approach under three distinct scenarios by rotating which of the three classes good, minor damaged, or damaged acts as the new class to be detected. The remaining two classes are subsequently treated as the known classes for training the LDA model. Our hypothesis test is structured as a multi-stage process, as illustrated in Figure \ref{fig:hypothesis_pipeline}.
\makebox[\textwidth][c]{
\includegraphics[width=1.5\textwidth]{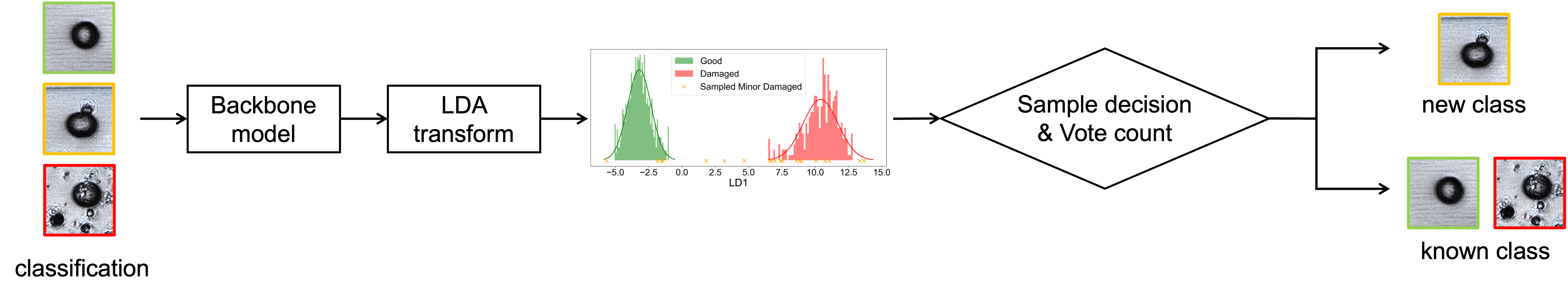}
}
\begin{figure}[H]
\centering

\caption{The multi-stage hypothesis testing pipeline. Features from the backbone are projected to 1D using LDA, then scored. A batch-level vote based on these scores determines if the batch belongs to a new class.}
\label{fig:hypothesis_pipeline}
\end{figure}

\subsubsection{ Dimensionality Reduction via LDA}

The 512 dimensional features extracted from the backbone model are first projected onto a lower dimensional manifold using LDA. The LDA model is trained only on the features from the two known classes after standardizing the data. We configure the LDA to find the single most discriminative component, effectively projecting all features onto a 1D line that best separates the two known distributions. All subsequent statistical analysis is performed in this 1D projected space.

\subsubsection{Novelty Scoring}
We define a novelty score $S(x)$ for any given sample $x$ based on its 1D projected value. We employ the minimum Mahalanobis distance to the known class means. This score measures the sample's distance, in terms of standard deviations, from the nearest known class center. Given the means ($\mu_1, \mu_2$) and the pooled variance ($\sigma_{\text{pooled}}^2$) of the two known classes on the 1D axis, the score is calculated as:$$S(x) = \min \left( \frac{(x - \mu_{\text{1}})^2}{\sigma_{\text{pooled}}^2}, \frac{(x - \mu_{\text{2}})^2}{\sigma_{\text{pooled}}^2} \right).$$A high score $S(x)$ indicates that the sample is an outlier relative to both of the known class distributions.

\subsubsection{Sample Thresholding and Vote Calibration}
Our test operates at two levels. First, we establish a sample level novelty threshold, $T_{\text{sample}}$. This threshold is determined non-parametrically by calculating the 95th percentile of the novelty scores $S(x)$ from all samples in the known classes. Any individual sample $x_i$ with $S(x_i) > T_{\text{sample}}$ is considered a potential novelty. Second, our hypothesis test is performed on a batch of $N=20$ samples. We define a trial level decision based on a voting mechanism. For a given batch, we count the number of samples $V$ that exceed $T_{\text{sample}}$. The entire batch is declared new class if this vote count $V$ exceeds a vote threshold, $T_{\text{vote}}$.

Instead of selecting an arbitrary value for $T_{\text{vote}}$, we perform a vote calibration analysis to find a statistically grounded threshold. This analysis involves simulating hundreds of trials, where each batch (20 samples) is drawn exclusively from the known classes. For each simulated trial, we count the number of votes $V$ (where a vote is cast if $S(x_i) > T_{\text{sample}}$) to generate an empirical distribution of $V$ under the null hypothesis. This allows us to map any potential $T_{\text{vote}}$ to the resulting misidentification rate for known batches. Our goal is to select the minimum $T_{\text{vote}}$ that controls this misidentification rate at a desired significance level $\alpha$ (e.g., $\alpha = 0.05$). This optimal threshold, $T_{\text{vote}}^*$, is formally defined as:$$T_{\text{vote}}^* = \min \left\{ t \in \{0, \dots, N\} \mid P(V > t \mid \text{Known Classes}) \le \alpha \right\}.$$This calibration ensures our batch-level decision is based on a quantifiable statistical confidence.

\subsection{Few-Shot Learning for Incremental Learning}

After establishing a baseline 2-class model trained on two known classes, our objective is to incrementally adapt this model to recognize a new class, using only a limited number of $K=20$ "few-shot" examples. This process is designed to be efficient and to mitigate catastrophic forgetting of the original classes.

\subsubsection{Incremental Update Strategy}
The incremental learning process begins by structurally modifying the model. The final 2-class linear classifier is discarded and replaced with a new, randomly initialized 3-class classifier head.

To fine-tune the model, we construct a balanced few-shot training dataset. This dataset comprises all $K=20$ available minor damaged samples, combined with $K=20$ randomly selected samples from the two known classes. This rehearsal or replay strategy ensures that the model is re-exposed to the original classes while learning the new one, which is a critical step in preventing the catastrophic forgetting of previously learned knowledge.

\subsubsection{Two-Stage Fine-Tuning}
The model is then fine-tuned on this new 3-class, $3K$ samples dataset using a two-stage process. First, in the head tuning stage, the ResNet-18 backbone and feature processor block are frozen. Only the parameters of the new 3-class linear classifier are trained. This stage is run for 15 epochs with a learning rate of $1 \times 10^{-3}$, allowing the new classifier to quickly align with the existing, powerful feature representations. Second, in the end to end tuning stage, the entire model is unfrozen. All parameters are fine-tuned end-to-end for an additional 15 epochs using a conservative learning rate of $5 \times 10^{-5}$. This step allows the deep features to gently adapt to accommodate the new unknown class, improving separation without drastically disrupting the existing two known classes feature clusters.

Throughout both fine-tuning stages, we continue to use the dual-view augmentation strategy and the $\mathcal{L}_{SCL}$ as described in Section \ref{subsec:backbone}. This ensures that the model's representations remain robust to scale and perspective changes as it learns the new class.

\subsection{Few-Shot Learning for Domain Adaption}\label{subsec:domain_adaption}
To bridge the domain gap between the hemisphere (source domain) and cube (target domain) datasets, we employ a few-shot domain adaptation technique. This approach is based on the DANN architecture, modified to incorporate a few-shot learning strategy using a small, labeled target set. The goal is to learn feature representations that are both discriminative for the quality classification task and invariant to the specific input domain.

\begin{figure}[H]
    \centering
    \includegraphics[width=1\linewidth]{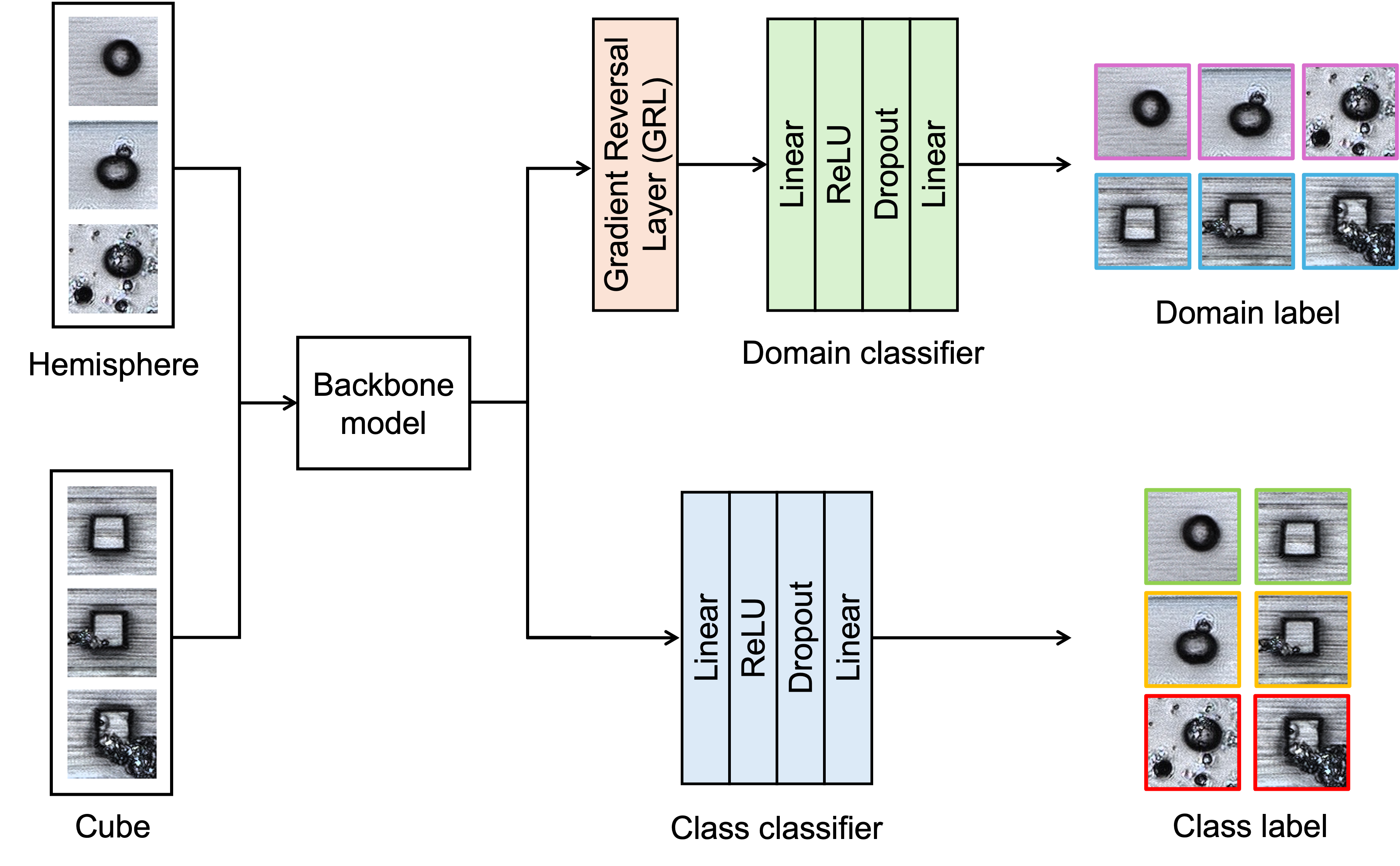}
    \caption{Architecture of the proposed DANN.}
    \label{fig:DANN}
\end{figure}

Our model is composed of three main components: a shared feature extractor, a class classifier, and a domain classifier as shwon in Figure \ref{fig:DANN}. The shared feature extractor utilizes the exact same backbone model described in Section \ref{subsec:backbone}, consisting of the truncated ResNet-18, SPP module, and feature processor, to produce a 512-dimensional feature vector, $z$. The class classifier takes these shared features $z$ as input to predict part quality, while the domain classifier takes the same features $z$ to predict the originating domain. The core of this method is a Gradient Reversal Layer (GRL) placed at the beginning of the domain classifier. The GRL acts as an identity function during the forward pass but reverses the gradient by multiplying it by a negative scalar, $-\lambda$, during backpropagation, enabling adversarial training.

This architecture is trained with a composite objective leveraging both the fully labeled source dataset and a few-shot target dataset (e.g., $K=5$ labeled samples per class). We define a Class Loss ($\mathcal{L}_{cls}$) as a weighted sum of supervised cross-entropy losses for both the source and target data:
\[
\mathcal{L}_{cls} = \mathcal{L}_{cls\_source} + w \cdot \mathcal{L}_{cls\_target},
\]
where $w=0.5$ in our implementation. This optimizes the feature extractor to be discriminative using all available label information. Simultaneously, a Domain Loss ($\mathcal{L}_{dom}$) is calculated by the domain classifier, which is a standard Cross-Entropy loss trained to distinguish source from target samples. The entire model is optimized on a total loss:
\[
\mathcal{L}_{total} = \mathcal{L}_{cls} + \lambda \cdot \mathcal{L}_{dom}.
\]
Due to the GRL, this adversarial process forces the shared extractor to learn a feature representation $z$ that is simultaneously class-discriminative (by minimizing $\mathcal{L}_{cls}$) and domain-invariant (by maximizing $\mathcal{L}_{dom}$).By trying to fool the domain classifier via the GRL, the feature extractor is encouraged to make hemisphere and cube features look similar. This strategy allows the model to learn a robust feature space that generalizes from the source to the target domain, even with minimal labeled target data.

This few-shot adversarial strategy allows the model to learn a robust feature space that generalizes from the source domain to the target domain, even with minimal labeled target data.

\section{Dataset} \label{sec:dataset} 
The visual characteristics of TPL-fabricated structures are critically influenced by the machine parameter setting (such as laser power, scanning rate), and machine condition. Variations in this machine setup can introduce distinct visual artifacts, such as noise patterns or defects creating a significant domain gap between datasets collected from different sources. This domain shift presents a major challenge for developing robust, generalizable quality inspection models.

Our dataset is composed of images from two distinct fabrication domains, hemisphere and cube. These datasets originate from two different TPL machine setups and exhibit a clear visual shift. Most notably, the cube domain images are characterized by prominent horizontal line artifacts, likely stemming from a machine-specific issue, which are not present in the hemisphere data.

\begin{figure}[H]
\centering
\makebox[\textwidth][c]{
\includegraphics[width=0.7\textwidth]{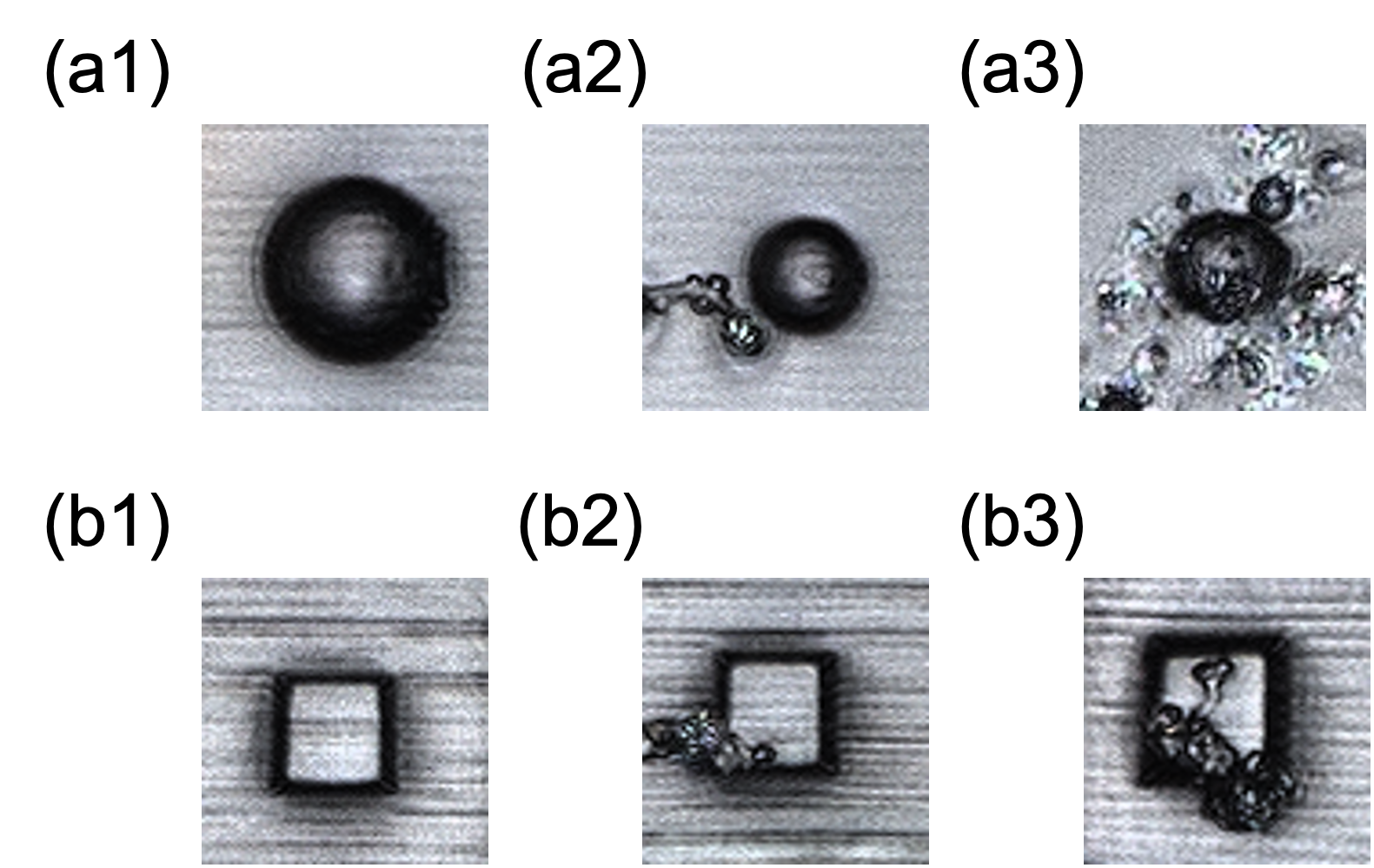}
}
\caption{Sample images from the two fabrication domains. Hemisphere domain: (a1) good; (a2) minor damaged; and (a3) damaged. Cube domain: (b1) good; (b2) minor damaged; and (b3) damaged. }
\label{fig:dataset}
\end{figure}

Within both domains, structures were categorized into three quality conditions based on the fabrication outcome: good, minor damaged, and damaged. Figure \ref{fig:dataset} illustrates these quality classes and highlights the visual domain gap between the hemisphere and cube image sets.

The distribution of images across these domains and quality conditions is summarized in Table \ref{tab:dataset_distribution}. The hemisphere dataset serves as our primary source domain, while the cube dataset, with its limited samples, represents a target domain for adaptation.

\begin{table}[h]
\caption{Distribution of Images in the Dataset}
\label{tab:dataset_distribution}
\centering
\begin{tabular}{lccc}
\hline
\textbf{Domain} & \textbf{Good} & \textbf{Minor Damaged} & \textbf{Damaged} \\
\hline
Hemisphere & 975 & 294 & 364 \\
Cube       & 179 & 21  & 11  \\
\hline
\end{tabular}
\end{table}

\section{Results and Discussion}\label{sec:results}
\subsection{Hypothesis Testing}
We evaluated the hypothesis testing approach under three distinct scenarios, where each of the three classes (good, minor damaged, damaged) took a turn acting as the new, unknown class to be detected.

\subsubsection{LDA Feature Space Separation} 
\begin{figure}[H]
\centering
\makebox[\textwidth][c]{
\includegraphics[width=0.7\textwidth]{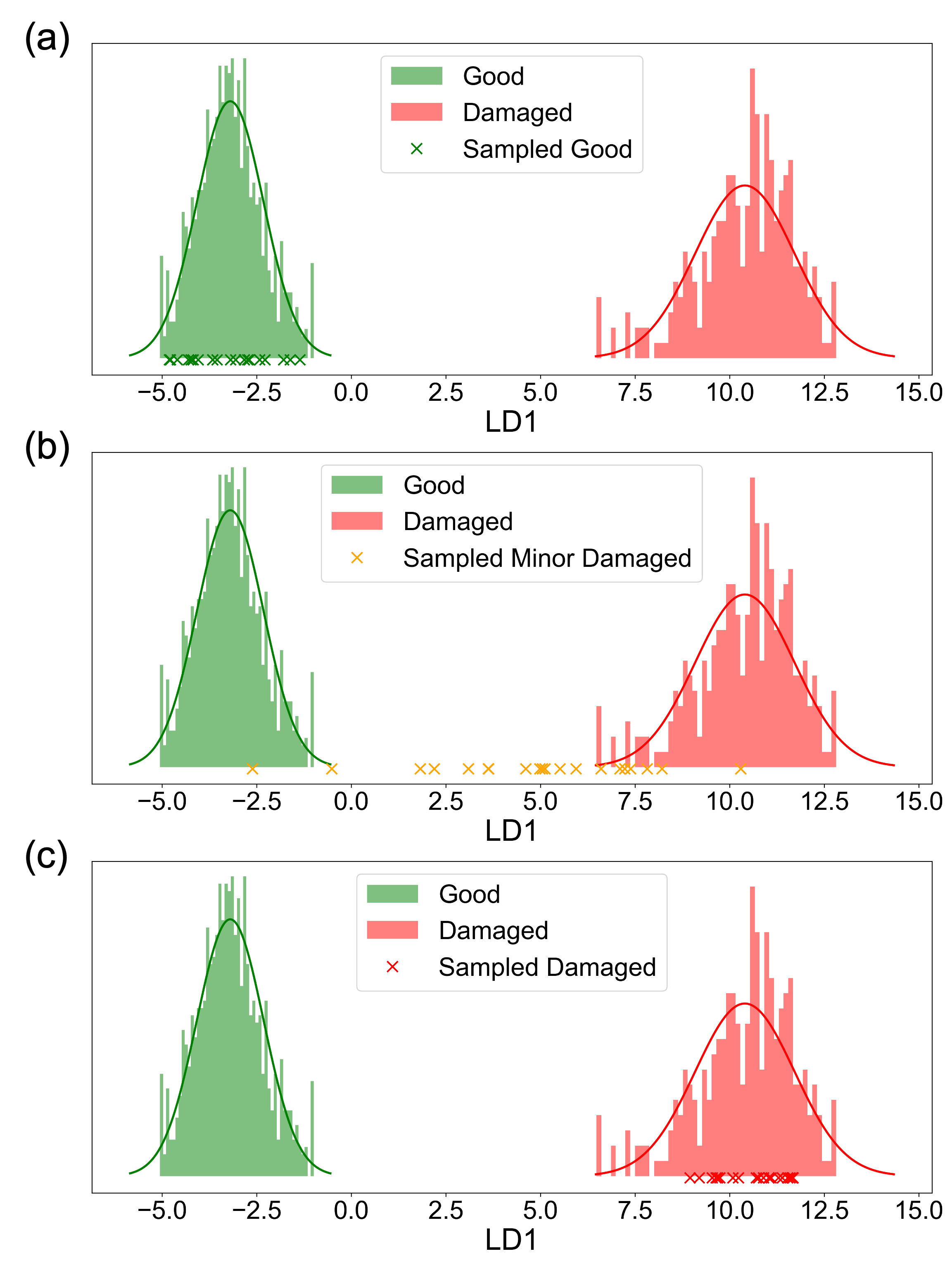}
}
\caption{1D LDA projections with minor damaged as the new class. (a) validation with known good samples; (b) novelty detection of unknown minor damaged samples; (c) validation with known damaged samples.}
\label{fig:lda_projections}
\end{figure}

The initial step was to train a 1D LDA on the two known classes (good and damaged) and project test class onto the same single axis (LD1). The results, visualized in Figure \ref{fig:lda_projections}, demonstrate the inherent separability of the classes. The LDA projection separates the two known classes, good and damaged, establishing a clear decision boundary. As a validation step, Figure \ref{fig:lda_projections}(a) and (c) demonstrate feature space stability, as new samples from these known classes correctly project onto their respective distributions. Crucially, when the unknown minor damaged samples are projected (Figure \ref{fig:lda_projections}(b)), they cluster in a unique location between the two known distributions. This indicates that the 1D LDA axis effectively captures a continuous range of fabrication quality or a damage level where minor defects naturally occupy an intermediate position between good and fully damaged states. Complete results for the other two scenarios are provided in \ref{app:lda_full}.

\subsubsection{Vote Threshold Calibration}
\begin{figure}[H] 
\centering  \includegraphics[width=0.7\linewidth]{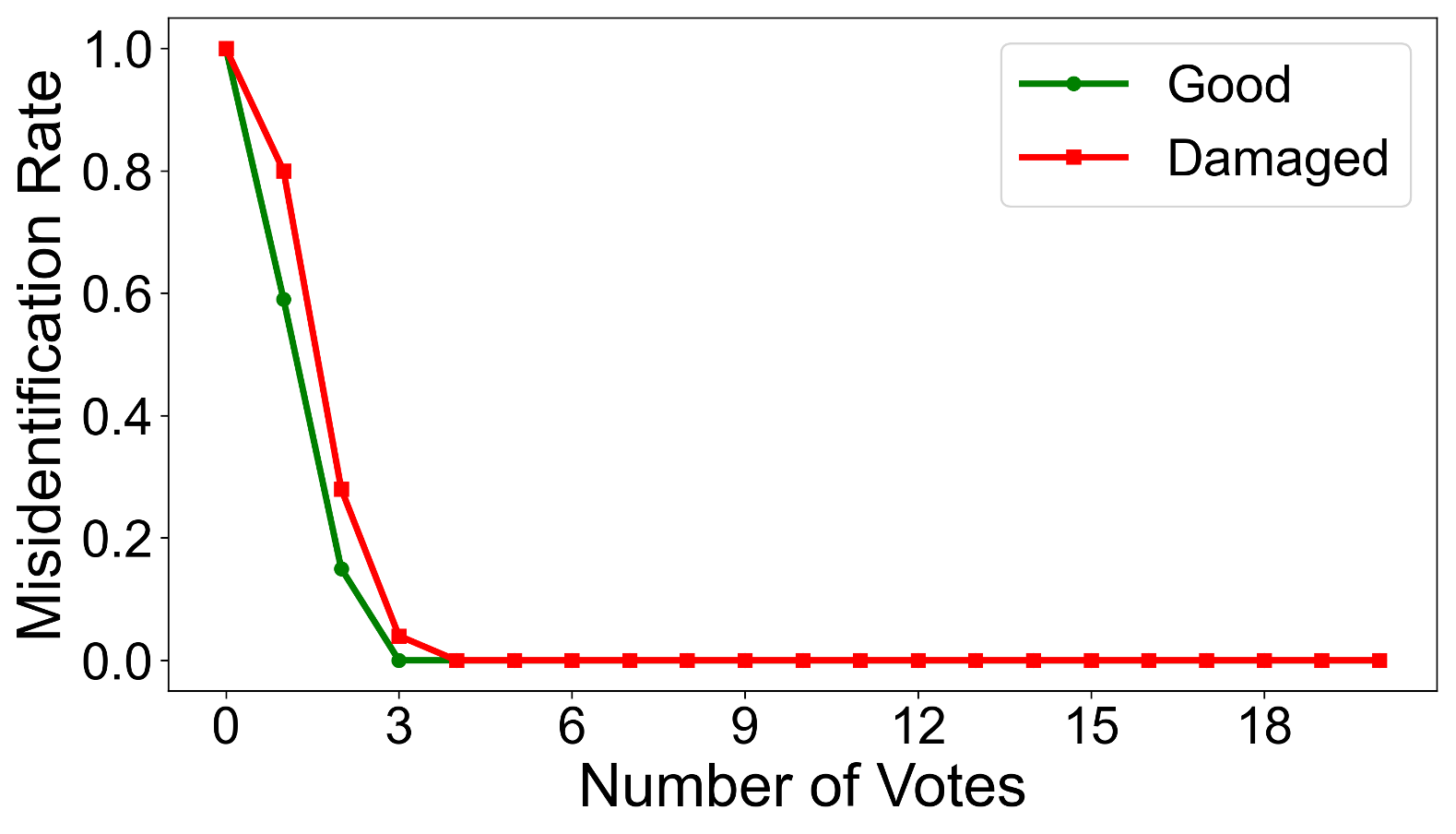} \caption{Vote calibration curves showing the misidentification rate versus the number of votes ($T_{vote}$) required for rejection. The plot corresponds to the scenario where good and damaged are the known classes, showing the rate at which batches from these classes are misidentified as new.}\label{fig:vote_calibration} \end{figure}

As described in the methodology, we performed a vote calibration to find a statistically robust vote threshold ($T_{\text{vote}}$) by analyzing the misidentification rate. Figure \ref{fig:vote_calibration} plots the misidentification rate as a function of the number of votes for the good and damaged as known classes scenario. The other two scenarios are shown in \ref{app:vote_calibration}.

The results are highly consistent across all three setups. The misidentification rate, which measures how often a known batch is falsely classified as new, drops rapidly as the vote threshold increases. In all cases, selecting a vote threshold of $T_{\text{vote}} = 4$ (i.e., requiring 4 or more samples in a batch of 20 to be considered as a new class) reduces the misidentification rate to near zero. This provides a strong statistical justification for using $T_{\text{vote}}=4$ as our decision boundary.

\subsubsection{Hypothesis Test Performance}
Using the established 95th percentile $T_{sample}$  and $T_{vote}=4$ from vote calibration, we conducted 100 trials for each class within each scenario to determine the final detection accuracy. The results are summarized in Table \ref{tab:hypothesis_results}. The table shows the percentage of batches that were flagged as new. The diagonal entries represent the correct detection rate for the new class, while the off diagonal entries represent the misidentification rate (i.e., the rate of incorrectly flagging a known batch as new).

\begin{table}[H]
\caption{Batch classification results for hypothesis testing. Each row defines the scenario which class was treated as new. Columns represent the true class of the tested batch. }
\label{tab:hypothesis_results}
\centering
\begin{tabular}{l ccc}
\toprule
 & \multicolumn{3}{c}{\textbf{True Class of Tested Batch}} \\
\cmidrule(lr){2-4}
\textbf{New Class Scenario} & \textbf{Good} & \textbf{Minor Damaged} & \textbf{Damaged} \\
\midrule
\textbf{Good} & 100\% & 0\% & 1\% \\
\textbf{Minor Damaged} & 0\% & 100\% & 8\% \\
\textbf{Damaged} & 0\% & 14\% & 99\% \\
\bottomrule
\end{tabular}
\end{table}

The results are good, with a correct detection rate of 99--100\% for all three scenarios. The approach identifies batches of the new class with high accuracy. The misidentification rate is also very low in most cases (0--1\%). The only notable exceptions are an 8\% misidentification rate when testing damaged batches in the minor damaged as new scenario, and a 14\% misidentification rate when testing minor damaged in the damaged as new scenario. This is consistent with the LDA plots (see Appendix \ref{app:lda_full}), which showed the minor damaged and damaged distributions are in closer proximity, leading to a higher chance of misidentification between these two specific classes. Overall, the statistical method is validated as a highly effective approach for detecting novel class batches.

\subsection{Few Shots Learning for Incremental Learning}

\subsubsection{Feature Space Visualization} 
\begin{figure}[H] 
\centering  
\makebox[\textwidth][c]{
\includegraphics[width=1.3\textwidth]{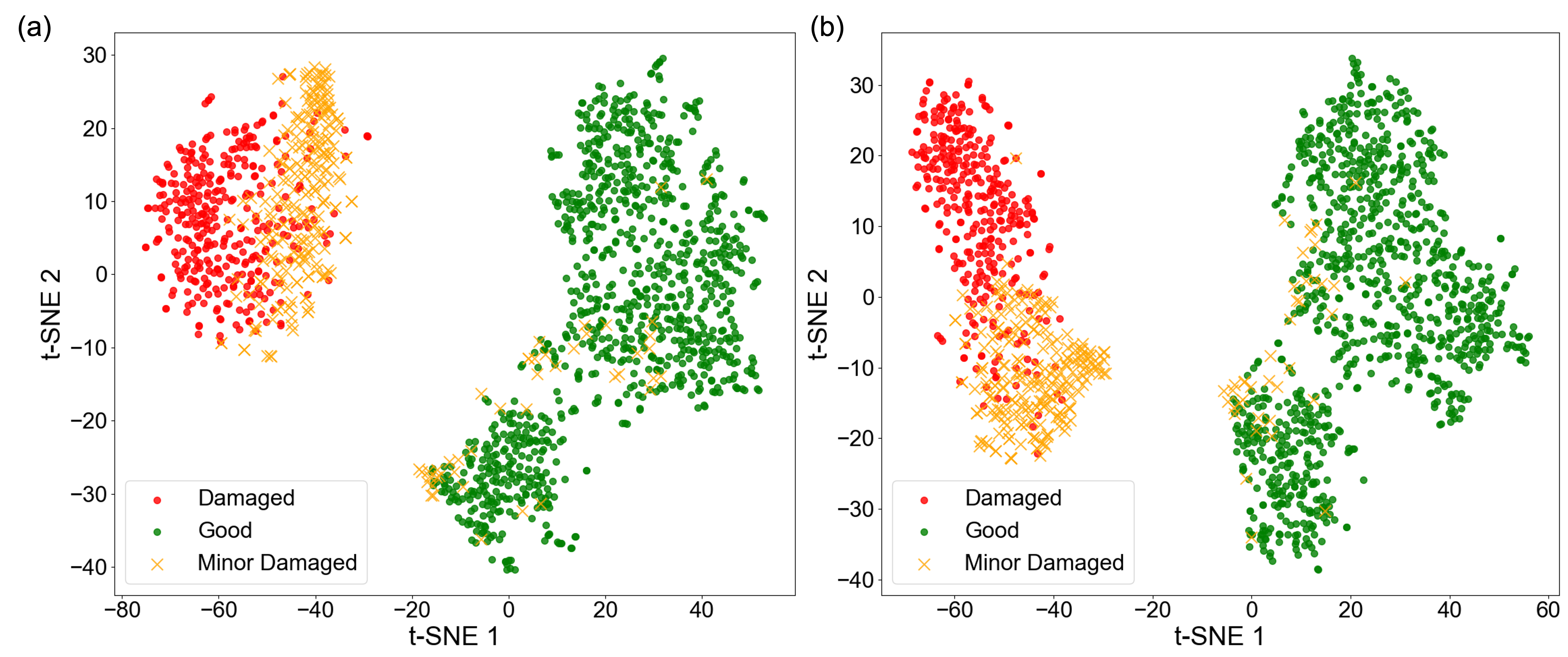}
}
\caption{t-SNE visualization of the feature space for minor damaged as a new class (a) before incremental learning and (b) after incremental learning}
\label{fig:tsne_incremental} \end{figure}

We first analyzed the impact of our incremental learning method on the feature space by visualizing it with t-SNE. Figure \ref{fig:tsne_incremental} shows the feature distributions of all three classes before and after the 20-shot fine-tuning for the minor damaged class as new class. Corresponding visualizations for the scenarios where good and damaged act as the new class are provided in \ref{app:tsne_additional}.

Initially, as shown in Figure \ref{fig:tsne_incremental}(a), the feature space is organized by the 2-class baseline model. The good samples in green color  form a well-defined cluster, separate from the damaged samples in red color. However, the unknown, minor damaged indicated as orange ``x'' samples are almost completely overlapped with the damaged cluster, indicating the baseline model is incapable of distinguishing them.

After applying our two-stage fine-tuning, a significant reorganization occurs, as seen in Figure \ref{fig:tsne_incremental}(b). The minor damaged samples have been pulled out from the damaged manifold to form their own distinct cluster, which now sits between the good and damaged classes. This qualitatively confirms that our method successfully learns a separable representation for the new class using only 20 samples, without causing the existing good and damaged clusters to collapse (i.e., without catastrophic forgetting).

\subsubsection{Comparative Performance vs. Baseline}
To quantify the benefit of our two-stage rehearsal strategy, we compared its performance against a baseline method. The baseline consists of a standard ResNet-18 fine-tuned directly on the three class imbalanced dataset. We repeated this experiment three times for each shot count (5, 10, 15, and 20) and for all three new class scenarios.

\begin{figure}[H] 
\centering \includegraphics[width=0.7\linewidth]{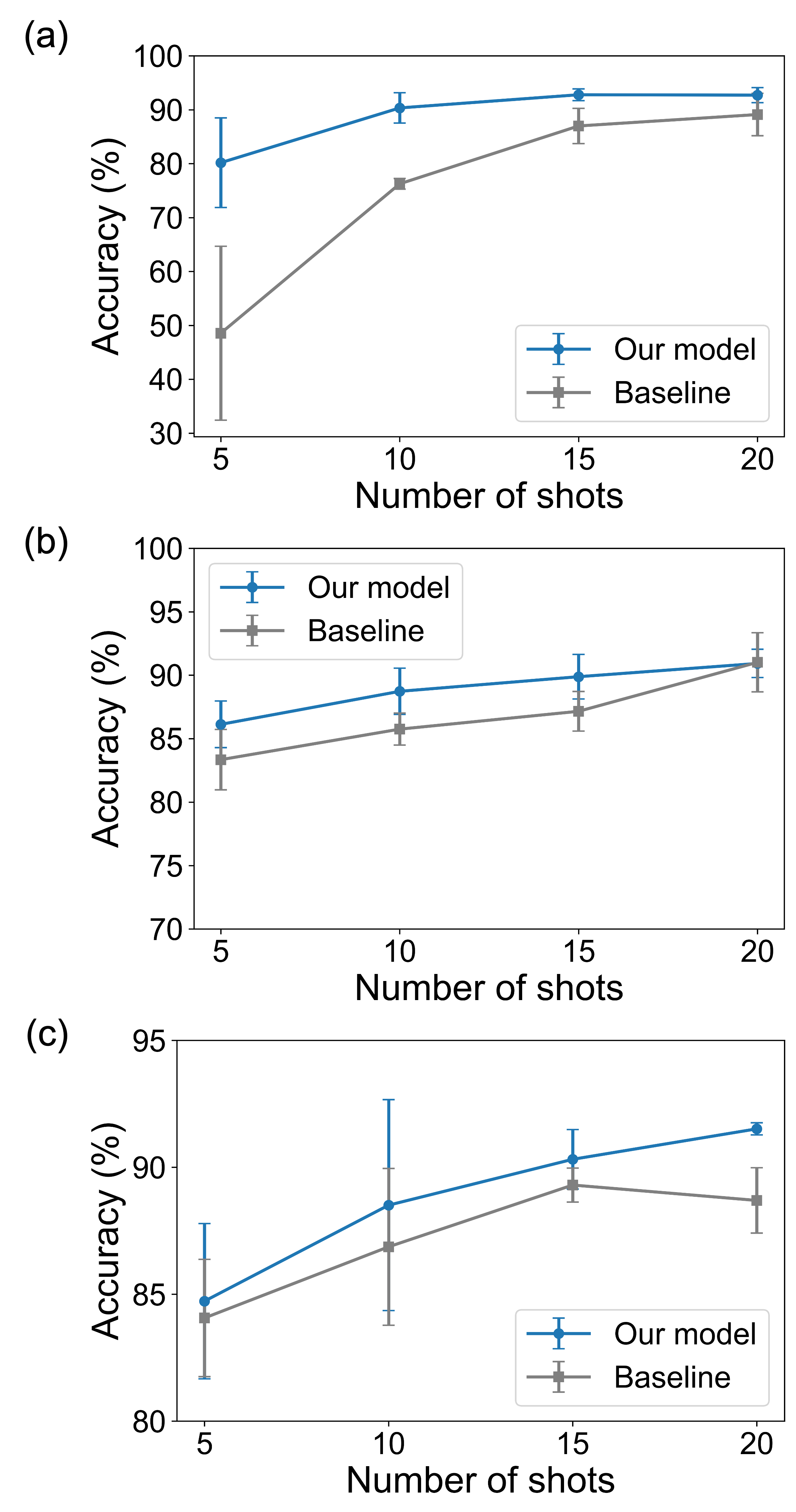} \caption{Accuracy comparison as a function of the number of shots (a) Good (b) Minor Damaged (c) Damaged as the new class} \label{fig:shots_compare} \end{figure}

The results, shown in Figure \ref{fig:shots_compare}, demonstrate the superiority of our two-stage approach over the baseline, especially in low data cases. This advantage is most pronounced when good is treated as the new class (Figure \ref{fig:shots_compare}(a)), where our model achieves ~80\% accuracy at only 5 shots, while the baseline struggles at ~50\%. When minor damaged is the new class (Figure \ref{fig:shots_compare}(b)), both models perform well due to the class's distinctness, but our model maintains a consistent lead. This advantage is also clear in the most challenging scenario where damaged is the new class (Figure \ref{fig:shots_compare}(c)). In this case, our model shows a widening accuracy gap over the baseline, achieving an average accuracy of over 91\% at K=20 shots compared to the baseline's 88.5\%.

\subsubsection{Incremental Learning Performance} 
\begin{figure}[H]
\centering
\makebox[\textwidth][c]{
\includegraphics[width=1.7\textwidth]{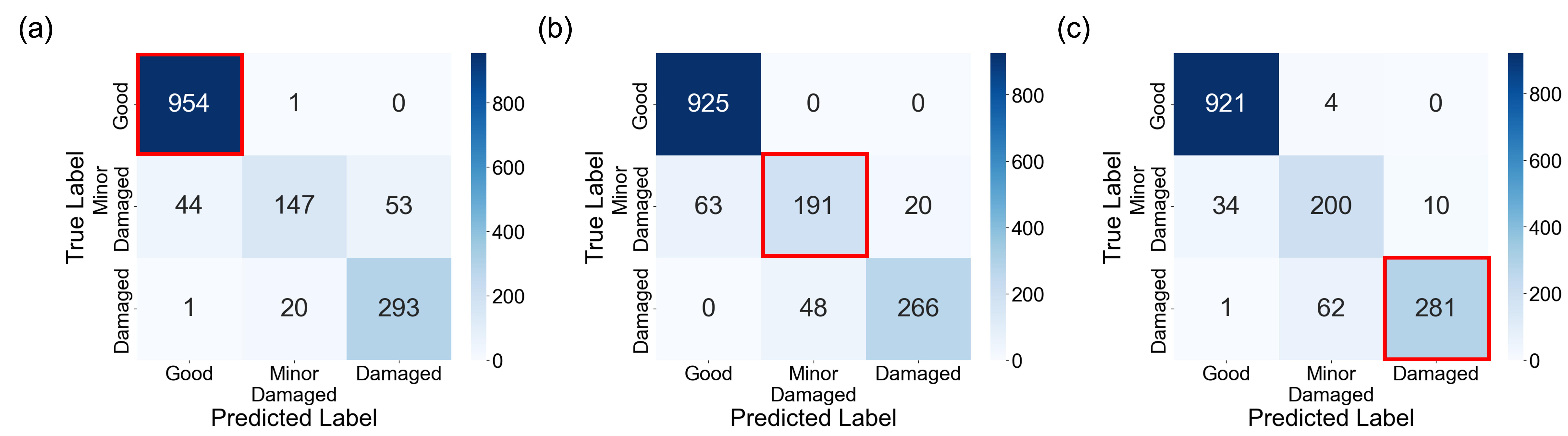}
}
\caption{Confusion matrices for the incremental learning test set. (a) Good; (b) Minor Damaged; (c) Damaged as the new class} 
\label{fig:conf_matrix} \end{figure}

We evaluated the performance of our two-stage incremental learning method on the full, unseen test set. The experiment was run three times, with each class (good, minor damaged, and damaged) serving as the new class to be learned from K=20 shots. The resulting confusion matrices are shown in Figure \ref{fig:conf_matrix}.

The model demonstrates strong overall performance, achieving 92.13\% accuracy with good as the new class, 91.34\% with minor damaged, and 92.66\% with damaged as shown in Figure \ref{fig:conf_matrix}.

The confusion matrices reveals specific performance characteristics for each scenario. When good is the new class (Figure \ref{fig:conf_matrix}(a)), the model learns it almost perfectly, achieving 99.9\% accuracy and demonstrating its ability to establish a clean decision boundary; errors in this case are mostly confined to the two original known classes. In contrast, when minor damaged is the new class (Figure \ref{fig:conf_matrix}(b)), the model faces its most difficult challenge. While the good class is perfectly preserved, the new minor damaged class is frequently misclassified as either good or damaged, which is consistent with its ambiguous position in the t-SNE visualization (Figure \ref{fig:tsne_incremental}). Finally, when damaged is the new class (Figure \ref{fig:conf_matrix}(c)), the model again preserves the good class and learns the new damaged class effectively. In this last scenario, the primary source of confusion is, as expected, between the adjacent damaged and minor damaged categories. This is expected due to the inherent visual similarity between these two classes, where the distinction can be subtle. This ambiguity may lead to potential inconsistencies in human labeling, as some images represent an intermediate state that is difficult to definitively categorize.

\subsection{Few Shots Learning for Domain Adaption}
\subsubsection{Feature Space Alignment} 
\begin{figure}[H] 
\centering \includegraphics[width=1\linewidth]{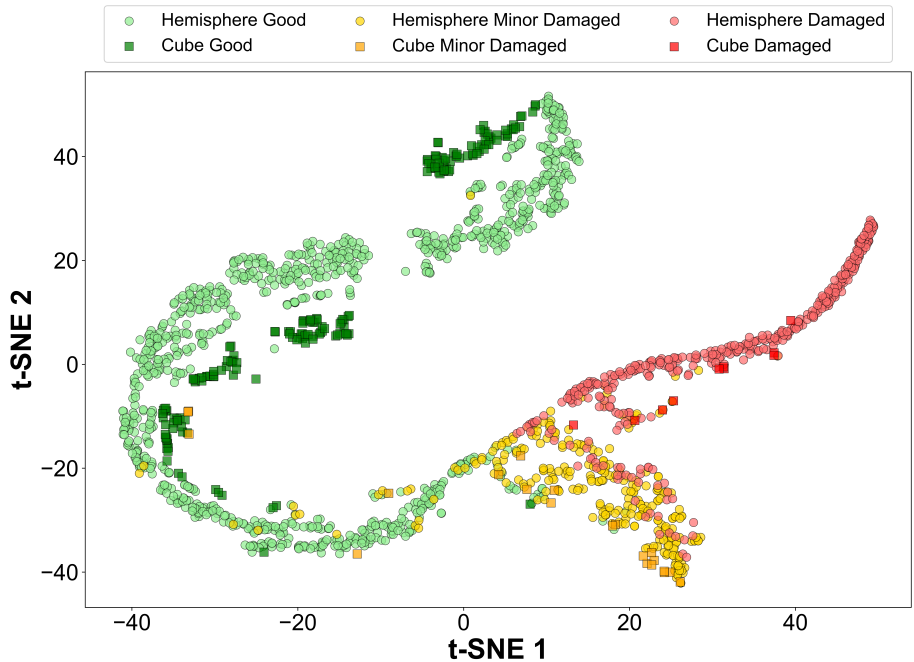} 
\caption{t-SNE visualization of the shared feature space after 5-shot domain adaptation.} \label{fig:tsne_domain} \end{figure}

We first visualized the 512-dimensional feature space of our Domain Adaption model after 5-shot adaptation using t-SNE, as shown in Figure \ref{fig:tsne_domain}. The plot reveals the effectiveness of the adversarial training. Despite the significant visual domain gap between the hemisphere and cube datasets (e.g., different design with the vertical line artifacts), samples from the same quality class are clustered together, irrespective of their domain.

For example, the cube good samples (green squares) are embedded within the larger hemisphere good cluster (green circles). Likewise, the cube minor damaged and cube damaged samples are successfully aligned with their hemisphere counterparts. This demonstrates that the shared feature extractor learned a domain-invariant representation, successfully aligning the cube target domain with the hemisphere source domain.

\subsubsection{Comparative Performance} 
To quantify the performance of our few-shot domain adaptation approach, we compared it against three baselines across 1 to 5 training shots from the target cube domain, with results summarized in Table \ref{tab:domain_comparison}. To ensure a fair comparison, all three baselines utilize the identical ResNet-18 backbone architecture. Baseline 1 represents a standard transfer learning approach, where the model is first fine-tuned on the 3-class source hemisphere domain and subsequently fine-tuned on the few-shot target cube data. Baseline 2 evaluates zero-shot transfer capability by taking the model fine-tuned on the source hemisphere domain and applying it directly to the target cube domain without further adaptation. Finally, Baseline 3 consists of a ResNet-18 fine-tuned exclusively on the limited 3-class samples from the target cube domain.

\begin{table}[H]
\caption{Comparison of mean accuracy (\%) with standard deviation (\%) on hemisphere and cube domains for 1--5 shots.}
\label{tab:domain_comparison}
\centering
\renewcommand{\arraystretch}{1.2}
\setlength{\tabcolsep}{6pt} 
\begin{tabular}{ll cccc} 
\toprule
\textbf{Shots} & \textbf{Domain} & \textbf{Baseline 1} & \textbf{Baseline 2} & \textbf{Baseline 3} & \textbf{Ours} \\
\midrule
\multirow{2}{*}{1 shot} & Hemisphere & $88.79 \pm 3.92$\%  & $92.99 \pm 0.30$\%  & $40.60 \pm 7.13$\%  & $88.96 \pm 2.56$\% \\
 & Cube & $58.41 \pm 20.82$\% & $13.49 \pm 2.81$\%  & $58.25 \pm 18.83$\% & $79.84 \pm 10.78$\% \\
\midrule
\multirow{2}{*}{2 shots} & Hemisphere & $90.46 \pm 1.50$\%  & $92.99 \pm 0.30$\%  & $56.62 \pm 20.39$\% & $92.58 \pm 0.50$\% \\
 & Cube & $64.60 \pm 14.55$\% & $13.49 \pm 2.81$\%  & $69.84 \pm 0.90$\%  & $88.73 \pm 3.89$\% \\
\midrule
\multirow{2}{*}{3 shots} & Hemisphere & $88.20 \pm 2.45$\%  & $92.99 \pm 0.30$\%  & $43.21 \pm 20.56$\% & $91.29 \pm 1.61$\% \\
 & Cube & $76.67 \pm 11.40$\% & $13.49 \pm 2.81$\%  & $63.17 \pm 21.77$\% & $92.22 \pm 2.21$\% \\
\midrule
\multirow{2}{*}{4 shots} & Hemisphere & $90.08 \pm 1.91$\%  & $92.99 \pm 0.30$\%  & $62.39 \pm 12.42$\% & $91.82 \pm 0.83$\% \\
 & Cube & $63.17 \pm 17.99$\% & $13.49 \pm 2.81$\%  & $89.84 \pm 7.32$\%  & $93.02 \pm 2.73$\% \\
\midrule
\multirow{2}{*}{5 shots} & Hemisphere & $90.10 \pm 1.35$\%  & $92.99 \pm 0.30$\%  & $55.08 \pm 24.47$\% & $91.21 \pm 0.67$\% \\
 & Cube & $77.14 \pm 0.67$\%  & $13.49 \pm 2.81$\%  & $76.67 \pm 9.93$\%  & $96.19 \pm 1.94$\% \\
\bottomrule
\end{tabular}
\end{table}

Our model consistently and significantly outperforms all baselines on the cube target domain across all shot counts. At K=5 shots, our model achieves a mean accuracy of 96.19\% $\pm$ 1.94\% on the target domain, surpassing the next-best baseline by nearly 20 percentage points. This highlights the superiority of adversarial alignment over a simple two-stage fine-tuning.

The performance of the other baselines confirms the difficulty of the task. Baseline 2's static 13.49\% accuracy demonstrates a total failure of zero-shot transfer, confirming the severity of the domain gap. Baseline 3, trained only on the sparse target data, also performs poorly, indicating that the few-shot samples alone are insufficient for generalization.

Importantly, our model also maintains high and stable performance on the hemisphere source domain (e.g., 91.21\% at 5 shots), indicating that the domain alignment process does not lead to catastrophic forgetting of the source task.

\subsubsection{Error Analysis} 
\begin{figure}[H] 
\centering \includegraphics[width=1\linewidth]{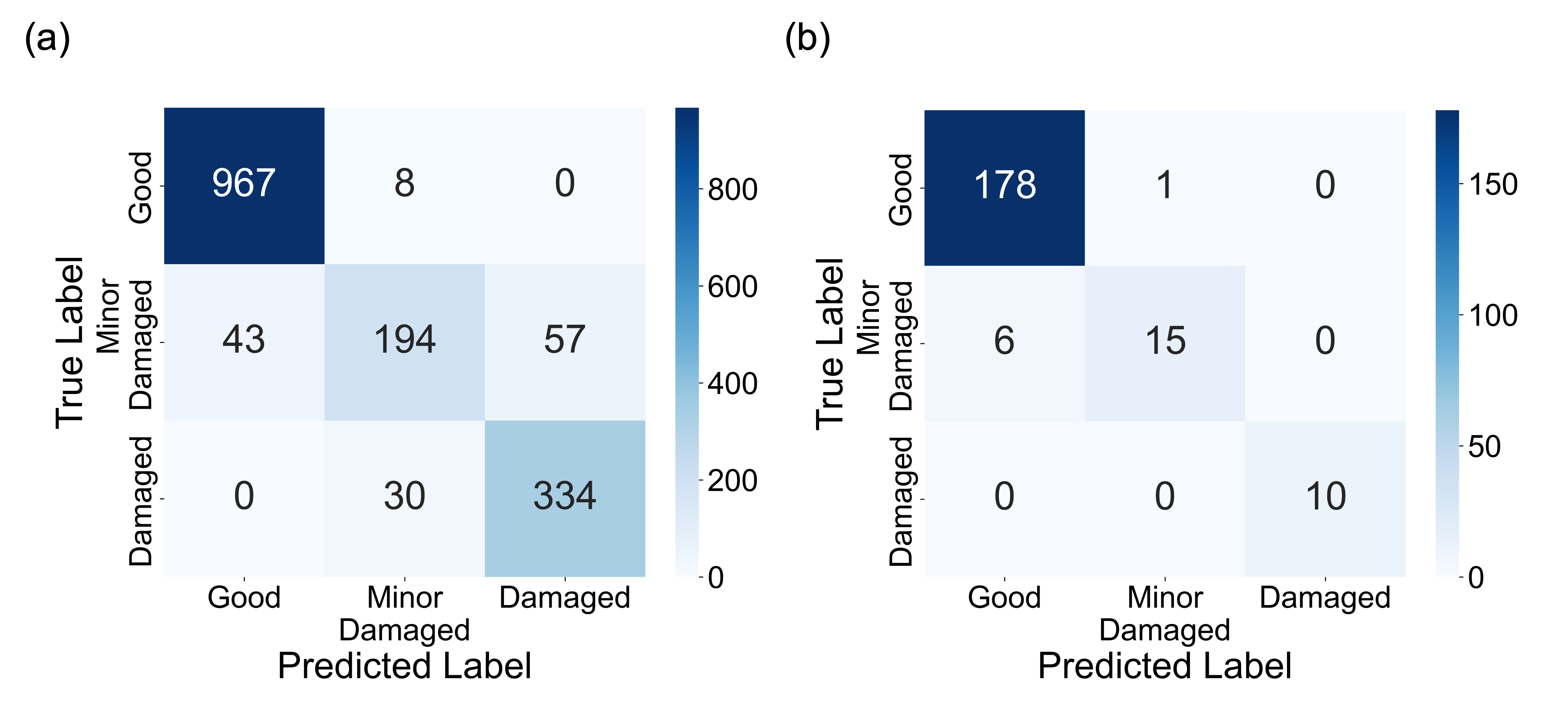} \caption{Confusion matrices from a single 5-shot experiment for (a) the hemisphere and (b) the cube} \label{fig:domain_confusion}  \end{figure}

Figure \ref{fig:domain_confusion} provides a detailed error analysis from a representative 5-shot run of our model, showing the confusion matrices for both the source hemisphere domain  and the target cube domain.

\begin{figure}[H] 
\centering \includegraphics[width=1\linewidth]{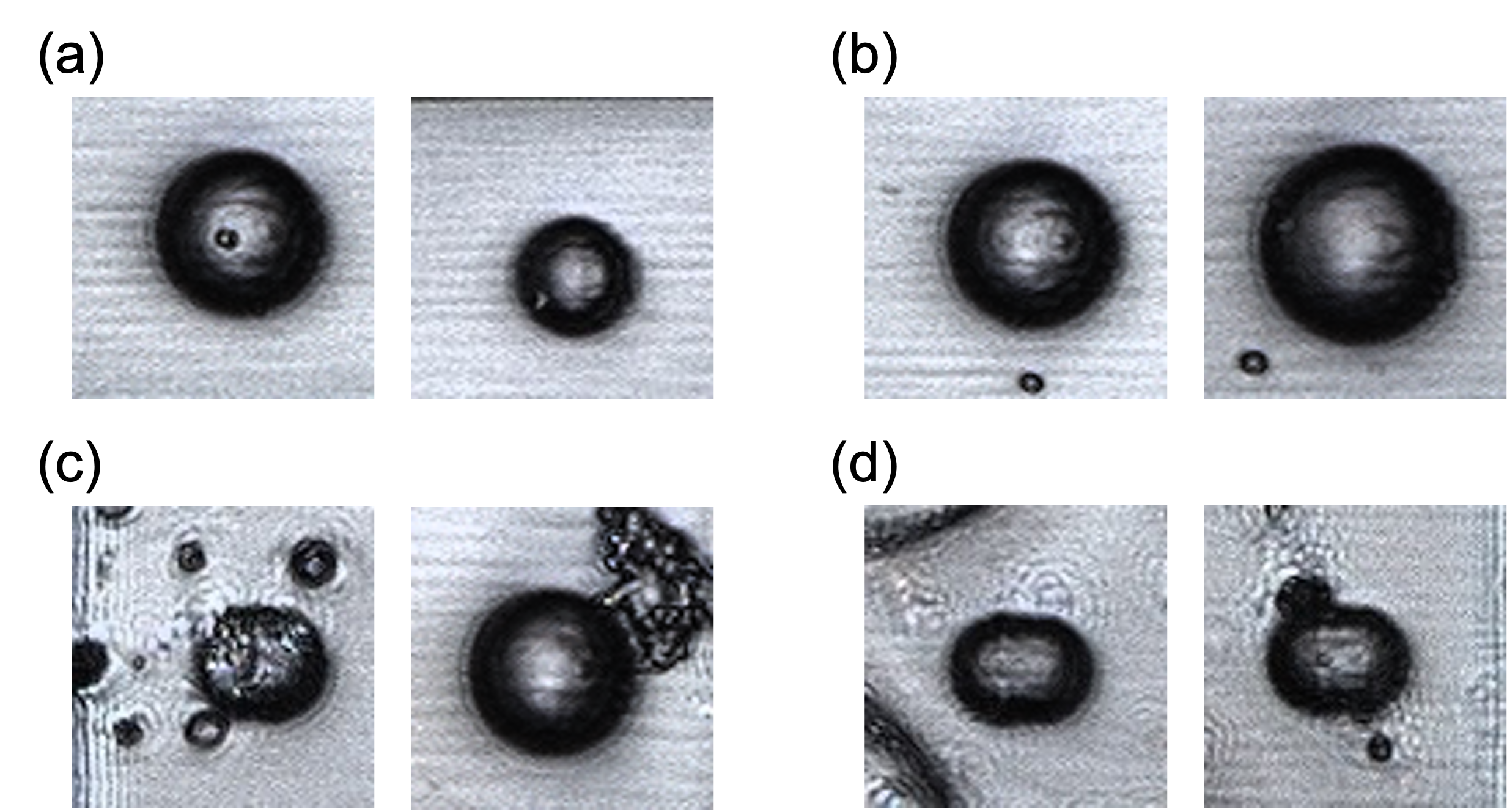} \caption{Misclassification examples from the hemisphere source domain: (a) True: Good, Predicted: Minor Damaged;
(b) True: Minor Damaged, Predicted: Good;
(c) True: Minor Damaged, Predicted: Damaged;
(d) True: Damaged, Predicted: Minor Damaged.} \label{fig:hemi_misclass}  \end{figure}

On the source domain (Figure \ref{fig:domain_confusion}(a)), this run achieves 91.55\% accuracy. The model remains highly effective at classifying the original good and damaged samples. The majority of errors are confined to the ambiguous minor damaged class. Figure \ref{fig:hemi_misclass} illustrates these specific misclassification scenarios, which highlight the difficulty in distinguishing visually similar classes. For instance, Figure \ref{fig:hemi_misclass}(a) shows good samples misclassified as minor damaged, likely due to a subtle surface artifact (e.g., a small circular detect) that the model mistakes for a defect. Conversely, Figure \ref{fig:hemi_misclass}(b) shows a minor damaged sample predicted as good, which occurs when the defect is small or distant from the main structure. The most significant confusion occurs between adjacent ambiguous classes: minor damaged predicted as damaged (Figure \ref{fig:hemi_misclass}(c)) and damaged predicted as minor damaged (Figure \ref{fig:hemi_misclass}(d)). These examples underscore a fundamental challenge: the actual defect level is a continuum, whereas the labels are discrete. This makes it inherently difficult, even for human annotators, to establish a clear and consistent decision boundary.

\begin{figure}[H] 
\centering \includegraphics[width=0.6\linewidth]{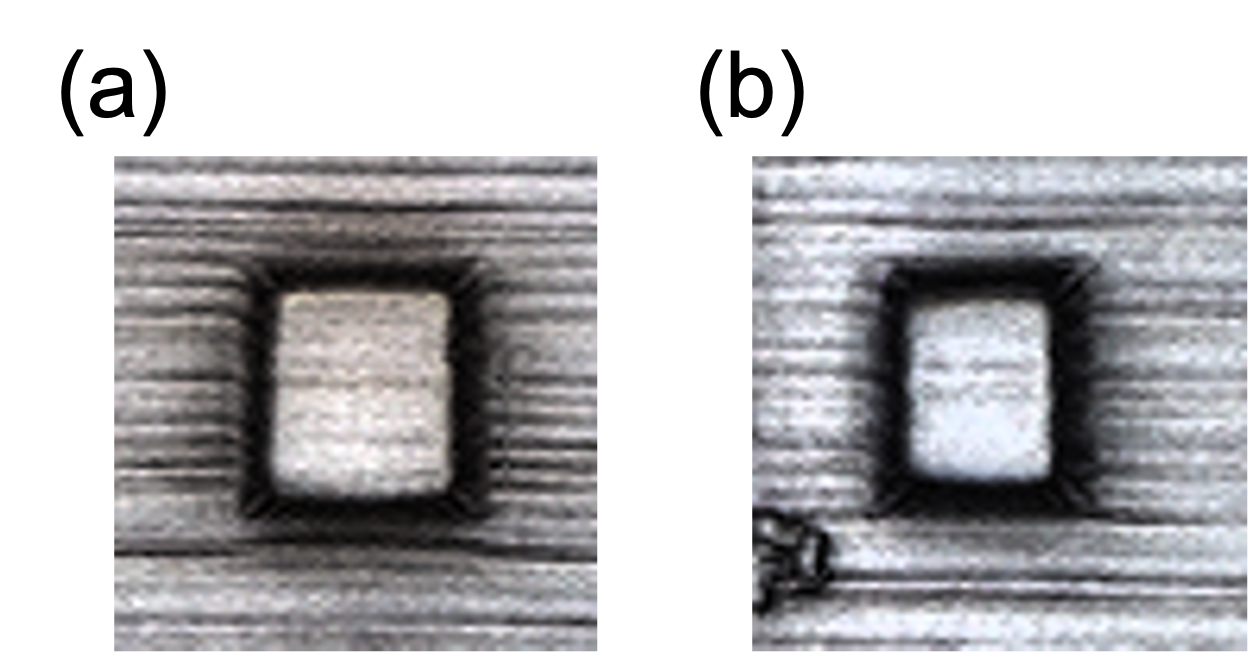} \caption{Misclassification examples from the cube target domain: (a) True: Good, Predicted: Minor Damaged;
(b) True: Minor Damaged, Predicted: Good.} \label{fig:cube_misclass}  \end{figure}

On the target cube domain (Figure \ref{fig:domain_confusion}(b)), this run achieves a 96.67\% accuracy. This single-run performance is consistent with the high average accuracy reported in Table \ref{tab:domain_comparison}. The model correctly classifies the majority of samples, including all 10 damaged samples. The only errors, illustrated in Figure \ref{fig:cube_misclass}, are the 7 misclassifications between the good and minor damaged classes. Specifically, 6 minor damaged samples were misclassified as good. Figure \ref{fig:cube_misclass}(b) shows a typical example of this, where the subtle defect on the sample's lower-left edge is heavily obscured by the horizontal line noise, causing the model to overlook it. The other error was a single good sample misclassified as minor damaged (Figure \ref{fig:cube_misclass}(a)). This may indicate the model is overly sensitive to the prominent dark border or noise artifacts, mistaking them for a defect. This performance on the target domain, using only 5 training shots per class, validates the effectiveness of our few-shot domain adaption approach.

\section{Conclusion and Future Work} \label{sec:conclusion}
In this paper, we present a computer vision framework designed to address critical, real-world challenges in quality inspection for TPL-fabricated structures. We introduced three core methodologies: a statistical hypothesis testing framework for novel class detection, an efficient two-stage incremental learning strategy for few-shot class addition, and a few-shot domain adaptation model for transferring knowledge to new part geometries. The effectiveness of these approaches was demonstrated on a challenging dataset of hemisphere and cube structures. Our hypothesis testing framework successfully identified new class batches with 99-100\% accuracy. The incremental learning method integrated a new class from only 20 samples to achieve ~92\% accuracy, significantly outperforming baseline fine-tuning. Furthermore, our domain adaptation model achieved 96.19\% accuracy on the target cube domain using only 5 labeled shots per class, overcoming a severe domain gap. These results demonstrate the framework's robustness, data efficiency, and adaptability. Overall, this research provides a comprehensive solution for deploying and maintaining robust quality inspection models in dynamic manufacturing environments where new defects and part types are continuously introduced.

Drawing on the findings of this work, several future research directions are identified. First, the current hypothesis testing framework operates at the batch level. Future work should focus on developing a more granular, sample level novelty detection system, perhaps by exploring reconstruction-based autoencoder methods or generative models to identify individual outlier samples in real-time. Second, our incremental learning model relies on a rehearsal buffer of old class samples. While effective, this may become cumbersome as the number of classes grows. Investigating rehearsal-free continual learning methods, such as those based on parameter regularization or dynamic network architectures, would be a valuable next step to improve scalability \cite{li2017learning,rebuffi2017icarl}. Finally, the domain adaptation was few-shot learning, requiring a few labeled target samples. A more advanced direction would be to explore unsupervised domain adaptation, which would align the domains without any target labels, or domain generalization, which aims to learn a single model that is inherently robust to domain shifts without seeing the target domain at all \cite{wang2022generalizing}.

\section*{Acknowledgments}

This research has been supported by the National Science Foundation, USA under Grant No. 2434813.

\appendix
\section{Hypothesis Testing Scenarios}
\label{app:lda_full}
\begin{figure}[H]
\centering
\includegraphics[width=0.8\textwidth]{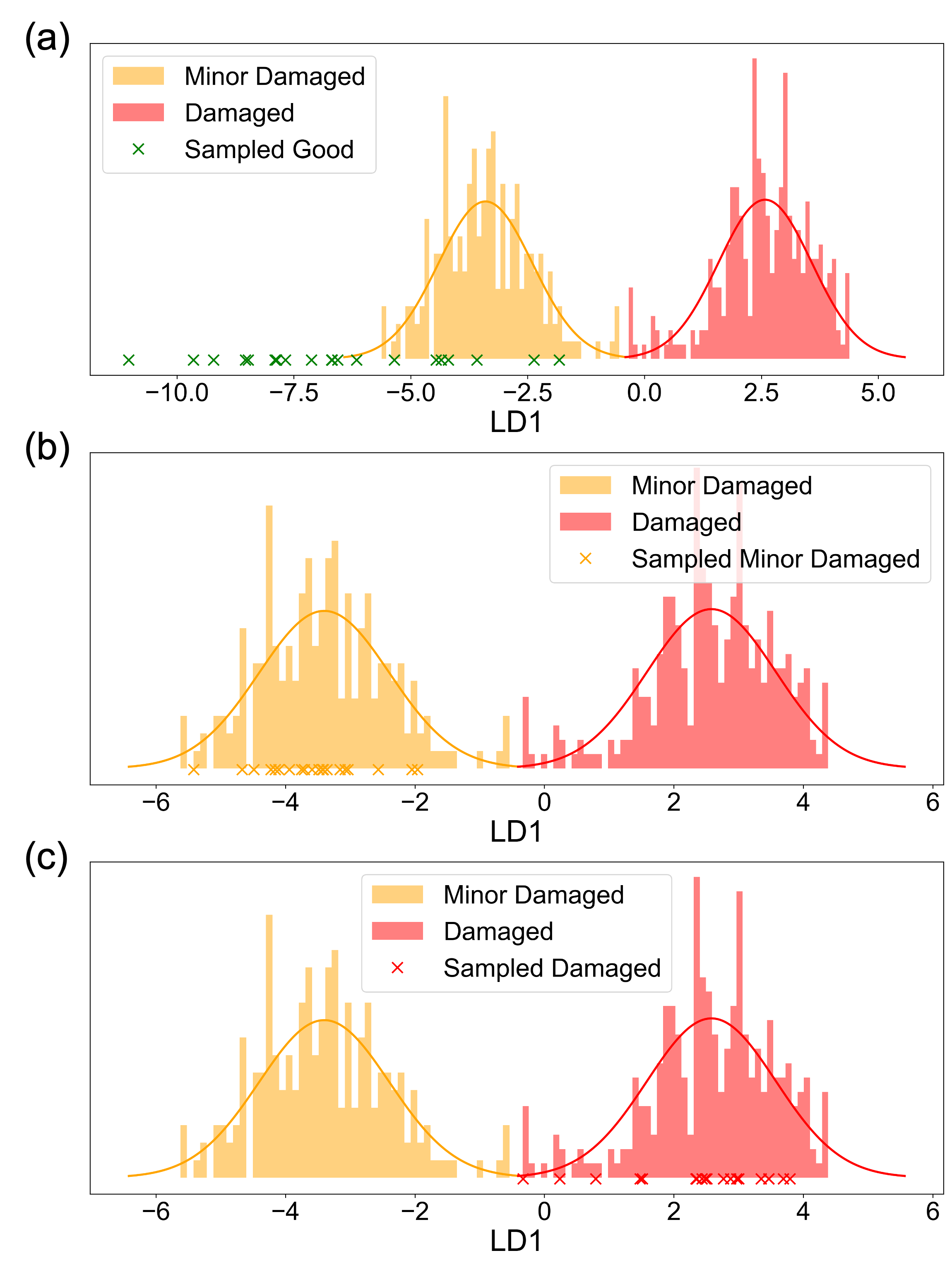}
\caption{1D LDA projections with good as the new class. (a) novelty detection of unknown good samples; (b) validation with known minor damaged samples; (c) validation with known damaged' samples.}
\label{fig:lda_good}
\end{figure}

\begin{figure}[H]
\centering
\includegraphics[width=0.8\textwidth]{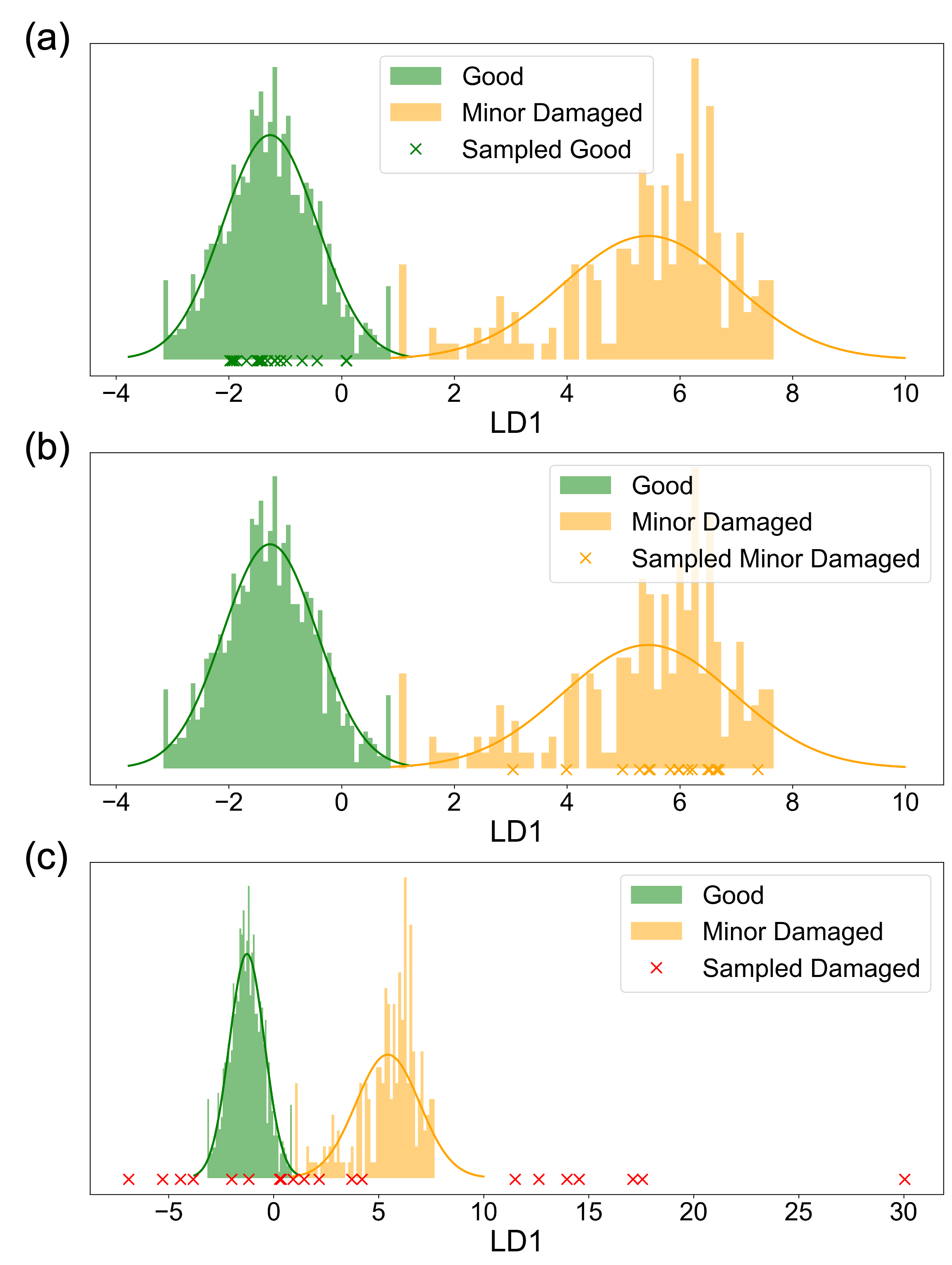}
\caption{1D LDA projections with damaged as the new class. (a) validation with known good samples; (b) validation with known minor damaged samples (c) novelty detection of unknown damaged samples.}
\label{fig:lda_damaged}
\end{figure}

This appendix presents the 1D LDA projection results for the two remaining experimental scenarios not covered in the main text: when good is treated as the unknown new class (Figure\ref{fig:lda_good}), and when damaged is treated as the unknown new class(Figure \ref{fig:lda_damaged}).

\section{Additional Vote Threshold Calibration Scenarios}
\label{app:vote_calibration}
As described in the main text, the vote threshold ($T_{\text{vote}}$) calibration was performed across three distinct scenarios to ensure robustness. The main text presented the results for the scenario where good and damaged classes were treated as known. This appendix presents the results for the other two scenarios.

\begin{figure}[H]
\centering
\includegraphics[width=0.7\linewidth]{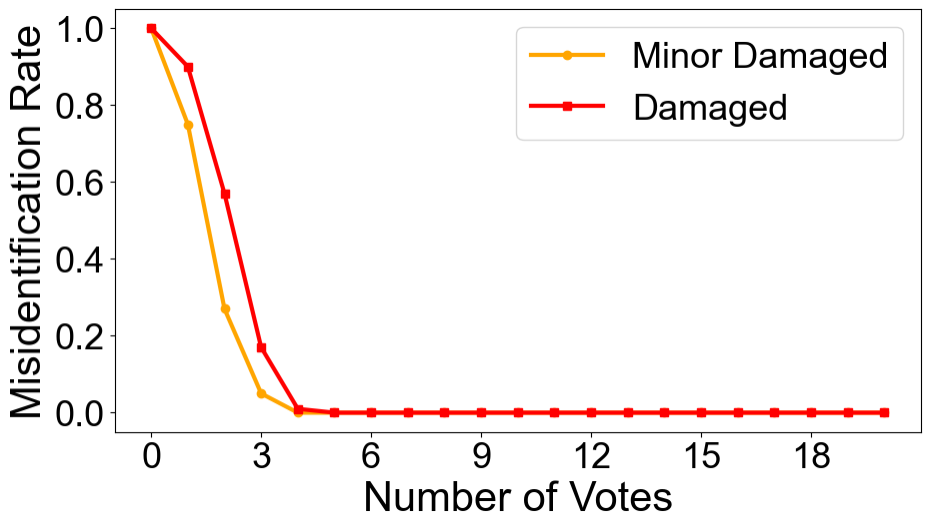}
\caption{Vote calibration curves showing the misidentification rate versus the number of votes ($T_{vote}$) required for rejection. The plot corresponds to the scenario where minor damaged and damaged are the known classes, showing the rate at which batches from these classes are misidentified as new.}\label{fig:vote_calibration_good_only}
\end{figure}

\begin{figure}[H]
\centering
\includegraphics[width=0.7\linewidth]{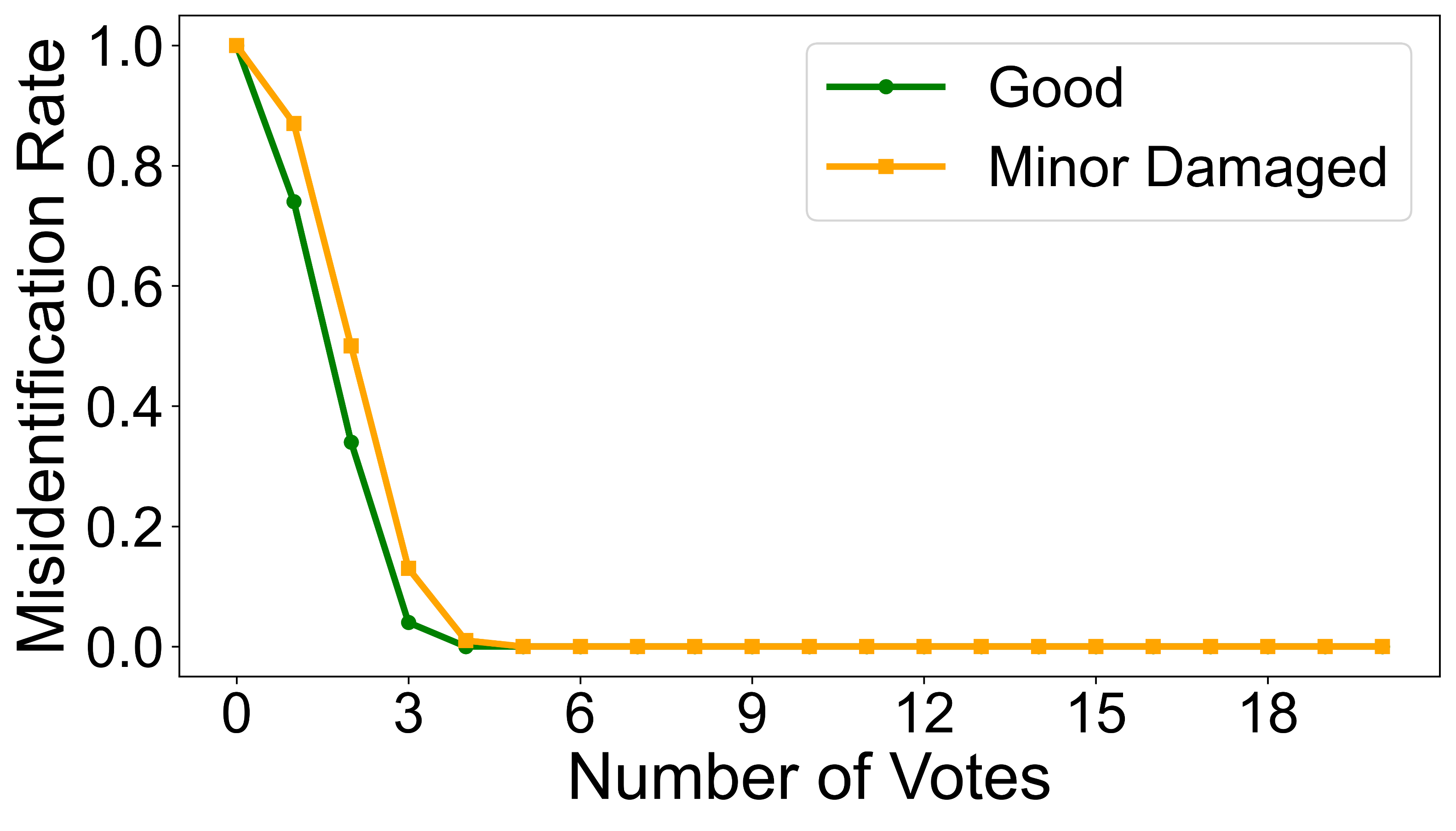}
\caption{Vote calibration curves showing the misidentification rate versus the number of votes ($T_{vote}$) required for rejection. The plot corresponds to the scenario where good and minor damaged are the known classes, showing the rate at which batches from these classes are misidentified as new.}\label{fig:vote_calibration_damaged_only}
\end{figure}

Figure \ref{fig:vote_calibration_good_only} and Figure \ref{fig:vote_calibration_damaged_only} shows the calibration curve that the results are highly consistent with the primary scenario. The misidentification rate rapidly approaches zero as the threshold increases.

\section{Feature Space Visualizations}
\label{app:tsne_additional}
\begin{figure}[H] 
\centering  
\makebox[\textwidth][c]{
\includegraphics[width=1.3\textwidth]{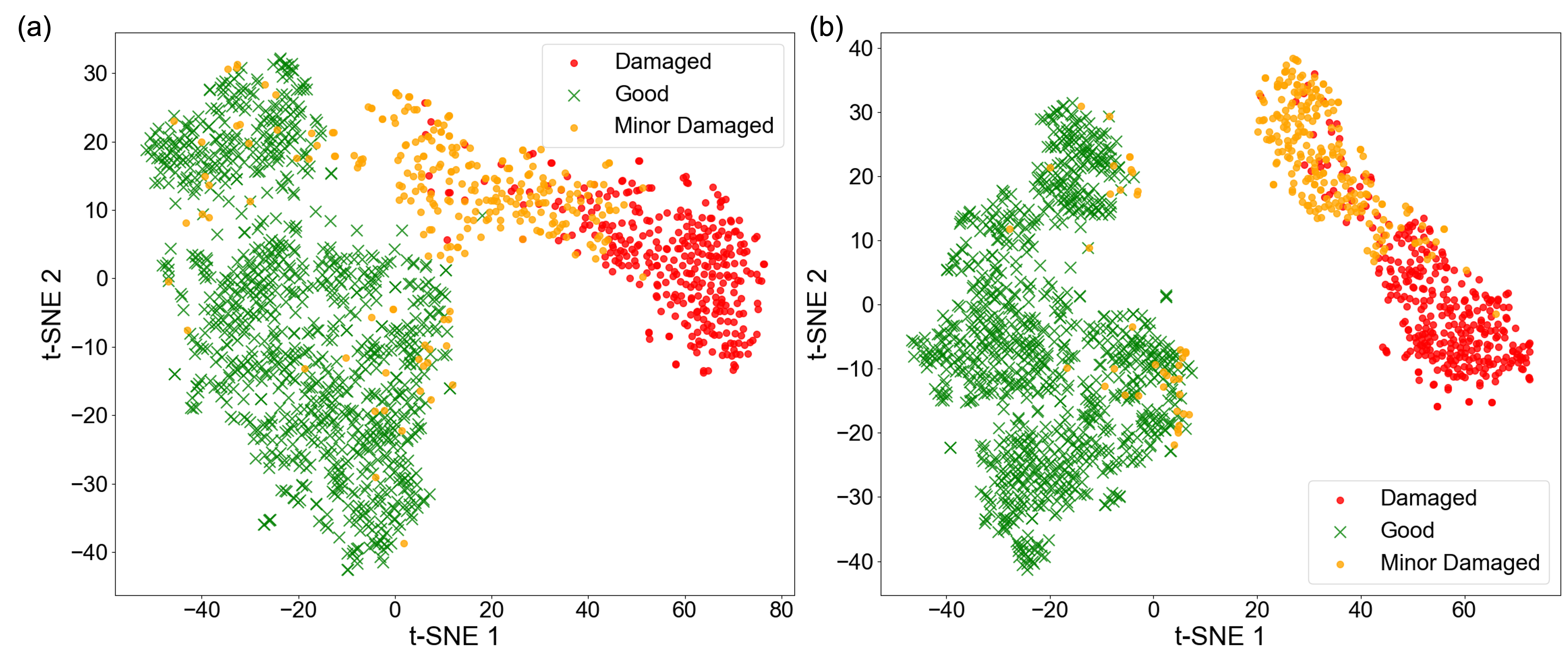}
}
\caption{t-SNE visualization of the feature space for good as a new class (a) before incremental learning and (b) after incremental learning}
\label{fig:tsne_good}
\end{figure}

\begin{figure}[H] 
\centering  
\makebox[\textwidth][c]{
\includegraphics[width=1.3\textwidth]{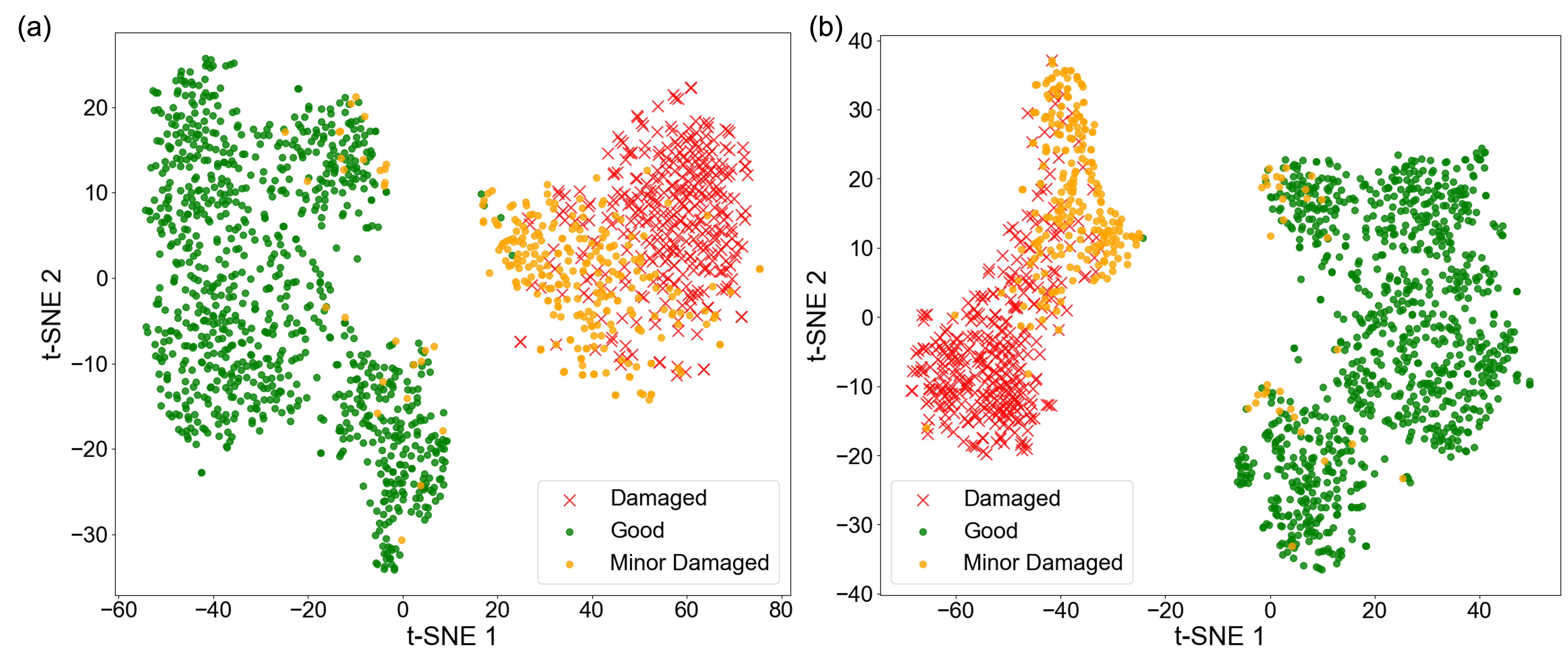}
}
\caption{t-SNE visualization of the feature space for damaged as a new class (a) before incremental learning and (b) after incremental learning}
\label{fig:tsne_damaged} \end{figure}

Figure \ref{fig:tsne_good} and Figure \ref{fig:tsne_damaged} presents the t-SNE feature space visualizations for the remaining two incremental learning scenarios.

\bibliography{ref}  
\end{document}